# Technical Correspondence

## Deep Models for Engagement Assessment With Scarce Label Information

Feng Li, Guangfan Zhang, Wei Wang, Roger Xu, Tom Schnell, Jonathan Wen, Frederic McKenzie, and Jiang Li

*Abstract*—Task engagement is defined as loadings on energetic arousal (affect), task motivation, and concentration (cognition) [1]. It is usually challenging and expensive to label cognitive state data, and traditional computational models trained with limited label information for engagement assessment do not perform well because of overfitting. In this paper, we proposed two deep models (i.e., a deep classifier and a deep autoencoder) for engagement assessment with scarce label information. We recruited 15 pilots to conduct a 4-h flight simulation from Seattle to Chicago and recorded their electroencephalograph (EEG) signals during the simulation. Experts carefully examined the EEG signals and labeled 20 min of the EEG data for each pilot. The EEG signals were preprocessed and power spectral features were extracted. The deep models were pretrained by the unlabeled data and were fine-tuned by a different proportion of the labeled data (top 1%, 3%, 5%, 10%, 15%, and 20%) to learn new representations for engagement assessment. The models were then tested on the remaining labeled data. We compared performances of the new data representations with the original EEG features for engagement assessment. Experimental results show that the representations learned by the deep models yielded better accuracies for the six scenarios (77.09%, 80.45%, 83.32%, 85.74%, 85.78%, and 86.52%), based on different proportions of the labeled data for training, as compared with the corresponding accuracies (62.73%, 67.19%, 73.38%, 79.18%, 81.47%, and 84.92%) achieved by the original EEG features. Deep models are effective for engagement assessment especially when less label information was used for training.

## I. INTRODUCTION

Assessment of human cognitive states such as vigilance, fatigue, and engagement has attracted considerable attention in recent years. Cognitive states can be evaluated by different approaches including subjective reports [2], biological measures (electroencephalography (EEG) [3], electrocardiogram (ECG) [4], electrooculography [5], and surface electromyogram [6]), physical measures (eye tracking [7], fixed gaze [8], and driving performance measures [9]), and hybrid methods [10]. However, correctly labeling these cognitive states is challenging and expensive. Conventional labeling methods can be categorized as indirect or direct methods [11], [12]. Indirect labeling methods usually evaluate cognitive levels in terms of task performances measured by metrics such as reaction time, error rate, etc. For complex tasks such as operating an aircraft or driving a car, these performance metrics are not easy to obtain. Direct labeling methods are manipulated by the involved subject through self-assessment, or by experts who have the domain knowledge. For complex and longtime tasks, self-assessment is either not feasible or can only provide a rough cognitive state estimate for the subject. In contrast, experts can provide more precise assessments by observing the subjects performance and considering the tasks phases and progress [13]. Although expert assessment is a relatively accurate labeling method for complex tasks, it usually provides scarce labeling data because large amounts of data are frequently in a middle or an unsure cognitive state and has to be discarded. In addition, manual examination of data is time consuming and expensive.

Most supervised algorithms suffer from the lack of sufficient training data because of the overfitting problem [14]. The problem is more severe if a model contains a large number of parameters. Unsupervised and semisupervised algorithms can utilize unlabeled data for training and thus improve the generalization capability of the model [15]–[17]. Deep learning can be considered as a semisupervised algorithm where both labeled and unlabeled data are used for training [18]. Deep Boltzmann machine (DBM) is a particular deep learning structure consisting of multiple layers of perceptrons for regression or classification [19], [20]. Although this deep structure is not new, training of the structure was not successful prior to 2006 because the training could easily get trapped in bad local minima [21]. In 2006, Hinton made a breakthrough in training the deep structure by utilizing the restricted Boltzmann machine (RBM) to effectively initialize the structure. He proposed the RBM algorithm to pretrain the deep structure in a layer-by-layer fashion using unlabeled data. The pretrained structure was then fine-tuned by labeled data using the backpropagation (BP) algorithm. A deep classifier for classification and an autoencoder for data reconstruction were designed based on the deep DBM structure and superior results were achieved [19], [20]. The training can be further improved by using dropout, which significantly reduces overfitting [22].

In this paper, we proposed two deep models, i.e., a deep classifier and a deep autoencoder, to tackle the challenge of the lack of label information for engagement assessment. The deep models were first pretrained with unlabeled data by unsupervised learning, and then, both were fine-tuned with limited label data. Before the fine-tuning by label information, the autoencoder was also optimized to reconstruct the unlabeled data. We evaluated the proposed models on 15 pilots who conducted a 4-h flight simulation from Seattle to Chicago, and their EEG signals were recorded during the simulation. Experts carefully examined the EEG signals and labeled 20 min of the EEG data for each pilot. The EEG signals were preprocessed and power spectrum features were extracted. These features were then input to the deep models to learn new representations for the EEG features. Finally, a linear support vector machine (SVM) was trained for engagement assessment.

The remainder of this paper is organized as follows. Section II describes our proposed methods. Section III presents the experimental setup, and Section IV illustrates results and discussions. Finally, Section V concludes the paper.

Manuscript received April 25, 2015; revised September 19, 2015, February 3, 2016, April 16, 2016, and July 21, 2016; accepted August 28, 2016. This work was supported by NASA under Grant NNX10CB27C. This paper was recommended by Associate Editor W. A. Chaovalitwongse.

F. Li, F. McKenzie, and J. Li are with the Department of Electrical and Computer Engineering, Old Dominion University, Norfolk, VA 23529 USA (e-mail: flixx003@odu.edu; rdmckenz@odu.edu; jli@odu.edu).

G. Zhang, W. Wang, and R. Xu are with Intelligent Automation, Inc., Rockville, MD 20855 USA (e-mail: guangfanzhang@gmail.com; wwllwei@gmail.com; hgxu@i-a-i.com).

T. Schnell is with the Department of Industrial Engineering, University of Iowa, Iowa City, IA 52242 USA (e-mail: thomas-schnell@uiowa.edu).

J. Wen is with the Department of Electrical and Computer Engineering, Georgia Institute of Technology, Atlanta, GA 30332 USA (e-mail: ancientwave11@gmail.com).

Color versions of one or more of the figures in this paper are available online at http://ieeexplore.ieee.org.

Digital Object Identifier 10.1109/THMS.2016.2608933

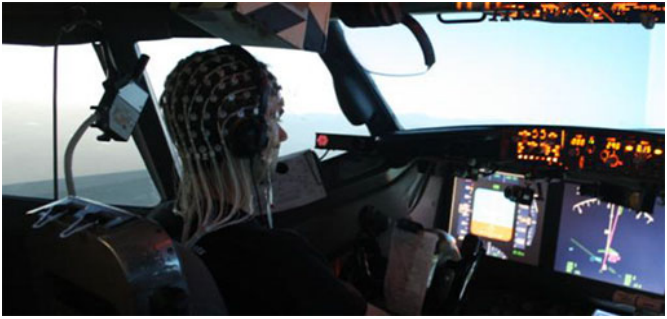

Fig. 1. Flight simulator.

## II. PROPOSED METHODS

In this section, we present the proposed engagement assessment system that consists of seven components including flight simulation and data collection, data labeling/ground truth definition, EEG data processing, feature extraction, deep learning models, deep classifier and deep autoencoder, and deep learning with dropout.

### A. Flight Simulation and Data Collection

A fully equipped Boeing 737 simulator was utilized in this study as shown in Fig. 1. A flight was simulated based on the details of an actual American Airlines flight from Seattle Tacoma international airport to Chicago O′Hare international airport. We invited 15 pilots to participate in this study. Each pilot has a commercial, a private, or an airline transport pilot license. To observe and measure pilot response to unexpected events, in addition to the simulated flight, three events were inserted: two air traffic control (ATC) calls and a failure event. The failure event was randomly selected from a hydraulic pump malfunction, wing body overheating, or window overheat, and all these failure events could be fixed by using instructions on a checklist.

The collected objective data include flight technical data (altitude, speed, etc.), physiological data (EEG, ECG, and eye tracking), audio, and video. The EEG sensor net was developed by ActiCap and contains 32 channels, which are located on the scalp as shown in Fig. 2. In addition, we collected several subjective evaluation data after each simulation, including situational awareness rating technique [23], Bedford workload scale [24], NASA Task Load Index (NASA TLX [25]), Samn-Perilli fatigue scale [26], and boredom proneness scale [27]. The performance data such as response times to ATC calls or failure events were recorded as well.

### B. Data Labeling/Ground Truth Definition

The engagement ground truth cannot be directly determined from any sensor. We proposed an approach to assess it by incorporating pilots' subjective evaluations, behavioral measurements, and physiological sensor measures [13].

The subjective evaluation data provided rough evaluations for engagement levels, which were collected after each pilot completed the flight simulation. The whole flight simulation was divided into 11 phases from takeoff to taxi and to gate. For each phase, each pilot gave self-assessment scores for workload scale, fatigue scale, and boredom proneness, which are correlated with engagement levels.

Behavioral measurements included communications with ATC and pilot response performance to expected/unexpected events. When an expected event happened, the pilots were usually awake resulting in increased engagement levels. When an unexpected event happened (we designed a pump failure of which the pilots were not aware in advance), the pilots had a more rapid engagement recovery and remained alert for a longer duration.

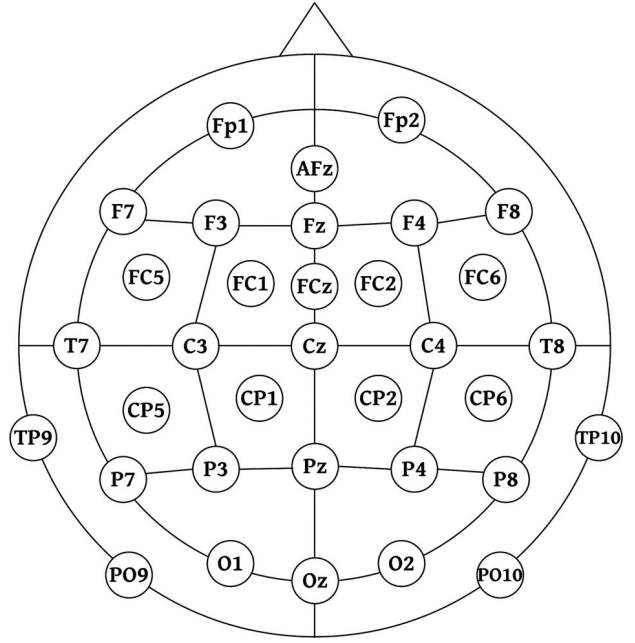

Fig. 2. EEG sensors.

TABLE I
PARAMETERS MONITORED FOR GROUND TRUTH DEFINITION

| Parameter Name | Category Name | Correlation |
| --- | --- | --- |
| Work load Fatigue scale Boredom proneness | Subjective evaluation | Referring to [23]–[27] |
| Unexpected event | Behavioral | Short response time → Engaged |
| Eye closure Head drooping R-R interval | Physiological | Low → Engaged |

We also utilized physiological sensor measures as strong disengagement/engagement indicators. We recognized eye closure/head drooping because of fatigue as an indicator of a disengaged state. The pilots' R-R (heart beat) interval was utilized as another indicator. High R-R interval values imply a relaxed stage during which engagement level degrades, and low R-R interval values indicate an engagement recovery stage. In total, 20 min of EEG recordings were labeled for each pilot, with 10 min recorded as engaged and another 10 min recorded as disengaged. Table I summarizes the parameters monitored for ground truth definition. More information can be found in [13].

### C. Electroencephalographic Data Processing

Our engagement assessment models used power spectrum features calculated from the collected EEG recordings. The literature suggested that some EEG channels were highly correlated with engagement and other cognitive states. Advanced brain monitoring (ABM) suggested bipolar sites Fz-POz and Cz-POz for engagement assessment, and C3-C4, Cz-POz, F3-Cz, F3-C4, Fz-C3, and Fz-POz for workload assessment [28]. A bipolar EEG signal is the potential difference between two EEG electrodes, which can be directly recorded if there is an EEG amplifier for each pair of electrodes. It can also be derived from unipolar measurements (e.g., $Potential_{C3-C4} = Potential_{C3} - Potential_{C4}$). Trejo *et al.* emphasized that mental fatigue was associated with Fz, P7, and P8 [29]. Pope *et al.* studied the combined powers of Cz, Pz, P3, and P4 for evaluating indices of operator engagement [30]. We started with all channels mentioned in these studies. Data collected from sensor Fz and P4 were not considered because their connections with the scalp were not good and led to heavy contaminations. Because the EEG sensors that were used (actiCAP) did not provide signals from channel POz, we selected Oz as a substitute, which is the nearest sensor to POz. To make it comparable, the sensors P7, P8, Pz, and P3 were paired with Oz, respectively. The final selected EEG sensors were Cz-Oz, C3-C4, F3-Cz, F3-C4, P7-Oz, P8-Oz, Pz-Oz, and P3-Oz.

EEG recordings are known to be contaminated by both physiological and nonphysiological artifacts [31]. In this paper, we first identified and removed spikes, amplifier saturations, and excursions by using the ABM model [28]. Second, we designed a high-pass filter with 0.5-Hz cutoff frequency to remove baseline drift, and a 60-Hz notch filter to delete the electrical interference. Finally, we implemented a wavelet-based method to remove physiological noises such as ocular and muscular artifacts [32]. The wavelet basis function was Coiflets, and the EEG data were decomposed using a six-level stationary wavelet transformation. Our EEG signals were sampled at 200 Hz, and the decomposed wavelet bands were 0–1.56, 1.56–3.13, 3.13–6.25, 6.25–12.5, 12.5–25, 25–50, and 50–100 Hz. For each wavelet band, the mean and standard deviation of the coefficients were calculated. Coefficients in the band were set to its mean if the absolute difference between the coefficient and the mean was larger than 1.5 times the standard deviation in that band. Finally, the EEG signals were reconstructed from the modified coefficients.

### D. Feature Extraction

A 3-s window was designed as an "epoch" and was shifted along the reconstructed EEG signals with a step size of 1 s, making a 2-s overlap between any two adjacent epochs. For each epoch, a 1-Hz frequency bin power spectral density (PSD) from 1 to 39 Hz for the selected eight EEG channels were calculated and yielded 312 ($39 \times 8$) 1-Hz bin PSDs as features. These feature vectors were then fed into the deep learning framework to learn a new feature representation for engagement assessment.

### E. Deep Learning Models

Deep models are capable of learning complex hierarchical nonlinear features, which are considered as better representations for original data in many fields such as speech recognition and computer vision [18]. However, the standard BP algorithm does not work well for deep structures with randomly initialized weights. Hinton *et al.* made deep architecture training possible by utilizing RBM to initialize the deep structure one layer at a time in an unsupervised fashion [19]. With unsupervised learning, deep learning tries to understand the training data first, i.e., to obtain a task specific representation from data so that a better classification can be achieved. It has experimentally proven that RBM plays a critical role in the success of deep learning [33].

RBMs are undirected graphical models that represent input data (visible units) $v \in R^D$ using binary latent variables (hidden units) $h \in \{0,1\}^K$. In this study, we used Gaussian–Bernoulli RBMs for training the first layer that contains real-valued visible units [34]. We utilized Bernoulli–Bernoulli RBMs for training higher layers that contain binary visible and hidden units. The joint probability distribution of $v$ and $h$ is

$$p(v,h) = \frac{1}{Z}\exp[-E(v,h)] \quad (1)$$

where $E$ is an energy function, and $Z$ is a normalization constant. The energy function for Gaussian RBMs is

$$E(v,h) = \frac{1}{2\sigma^2}\sum_i v_i^2 - \frac{1}{\sigma^2}\left(\sum_i c_i v_i + \sum_j b_j h_j + \sum_{i,j} v_i w_{ij} h_j\right) \quad (2)$$

where $c \in R^D$ and $b \in R^K$ are the biases for visible and hidden units, respectively, $W \in R^{D \times K}$ are weights between visible and hidden units, and $\sigma$ is the standard deviation associated with a Gaussian visible neuron $v_i$. The conditional probability distributions of the Gaussian RBM are as follows:

$$P(h_j = 1|v) = \text{sigmoid}\left(\frac{1}{\sigma^2}\left(\sum_i w_{ij} v_i + b_j\right)\right) \quad (3)$$

$$P(v_i|h) = N\left(\sum_j w_{ij} h_j + c_i, \sigma^2\right). \quad (4)$$

For Bernoulli–Bernoulli RBMs, the energy function and conditional probability distributions are as follows:

$$E(v,h) = -\left(\sum_i c_i v_i + \sum_j b_j h_j + \sum_{i,j} v_i w_{ij} h_j\right) \quad (5)$$

$$P(h_j = 1|v) = \text{sigmoid}\left(\sum_i w_{ij} v_i + b_j\right) \quad (6)$$

$$P(v_i = 1|h) = \text{sigmoid}\left(\sum_j w_{ij} h_j + c_i\right). \quad (7)$$

The parameters $W$, $b$, and $c$ of RBMs are learned using contrastive divergence [19]. For Gaussian–Bernoulli RBM, the formulas for updating those parameters during every iteration are

$$\Delta w_{ij}^{(n+1)} = \eta \cdot \Delta w_i j^{(n)} - \epsilon\left(<\frac{1}{\sigma^2} v_i h_j>_d - <\frac{1}{\sigma^2} v_i h_j>_m\right) \quad (8)$$

$$\Delta b_i^{(n+1)} = \eta \cdot \Delta b_i^{(n)} - \epsilon\left(<\frac{1}{\sigma^2} v_i>_d - <\frac{1}{\sigma^2} v_i>_m\right) \quad (9)$$

$$\Delta c_j^{(n+1)} = \eta \cdot \Delta c_j^{(n)} - \epsilon\left(<h_j>_d - <h_j>_m\right) \quad (10)$$

where $<\cdot>_d$ and $<\cdot>_m$ denote the expectation computed over data and model distributions accordingly, $n$ is the index for iteration, $\eta$ is the momentum, and $\epsilon$ is the learning rate.

### F. Deep Classifier and Deep Autoencoder

After pretraining, two types of deep models were constructed by adding a label layer (Type I network, deep classifier) or unfolding the pretrained model (Type II network, deep autoencoder) as shown in Fig. 3. Both models were initialized with the learned parameters ($W$ and $b$) and fine-tuned using the BP algorithm [20]. The layer under the label layer in the Type I network and the middle layer in the Type II network were new representations for the original features shown in yellow in Fig. 3. An SVM classifier was then trained using the learned representations for engagement assessment.

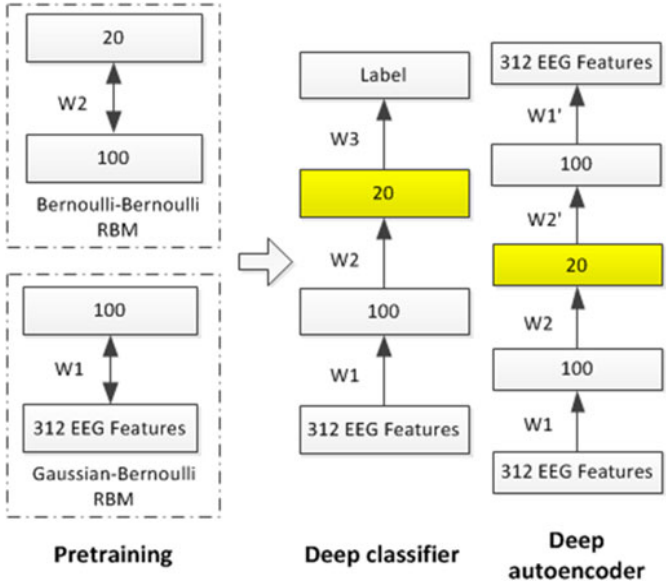

Fig. 3. Deep learning models.

### G. Deep Learning With Dropout

The dropout technique was proposed to reduce overfitting by preventing complex weights coadaptations [22]. During training at each iteration, a hidden unit was randomly omitted or dropped out from the network with a probability of $p$ (usually $p = 0.5$), which can decrease the correlations among different hidden units. Dropout can also be applied to the visible layer, and a common dropout probability for the visible layer is 0.2. During testing, the weights of the learned model should be scaled as $(1 - p)W$ to compensate for the dropout step in training. It has been shown that the dropout procedure performs similar to the $L_1$ or $L_2$ regularization, and it is especially useful when available data are limited [22], [35], [36].

## III. EXPERIMENTAL SETUP

### A. Data Preparation

The proposed system was evaluated on 15 pilots who each conducted a 4-h flight simulation. Experts labeled 10 min of EEG data as engaged and another 10 min of EEG data as disengaged [13] for each pilot. The EEG signals were first preprocessed including smoothing and artifacts removal. Then, the feature extraction procedure produced about 1200 PSD feature vectors for each pilot. Those feature vectors were then fed into the deep learning framework to learn new feature representations for the original features. Finally, the new feature representations were used for engagement assessment using a linear SVM classifier.

### B. Engagement Assessment Through Fivefold Cross Validation

We first conducted a fivefold cross validation (CV) on the labeled data from each of the 15 pilots to study engagement assessment. In fivefold CV, we randomly divided the labeled data into five parts. We first used four parts to train a model, and the trained model was then applied to the remaining part for evaluation. This procedure was repeated five times so that each part was tested once. The fivefold CV was done for each pilot separately so that each pilot had an individual model.

There were several hyperparameters associated with the deep learning scheme including the number of hidden layers in the deep structure and the number of hidden units in each hidden layer. Because of the computational complexity of deep learning and the limited labeled data, it is difficult to determine the optimal values for the hyperparameters in the deep learning structure. Instead, we studied the effects of these hyperparameters by performing fivefold CV on the available data with different combinations of parameter values and compared performance results from these combinations. In addition, we studied if dropout can improve performance of deep learning for engagement assessment.

### C. Engagement Assessment With Scarce Label Information

In the fivefold CV evaluation, we randomly divided the labeled data into five parts without considering the time information associated with each of the data points. In practice, a trained model may be applied for a certain amount of time without retraining, making the testing data continuous in time. To mimic the practical application scenario, we conducted a more strict engagement assessment for pilots using the limited labeled data. We designed experiments in which we utilized the continuous top 1%, 3%, 5%, 10%, 15%, and 20% of all labeled data, respectively, for training and the corresponding remaining data for testing. In each experiment, we first used all data points (without labels) to pretrain the deep structure. After pretraining, we fine-tuned the deep structure utilizing labels from the training data or by using the autoencoder. Finally, the learned 20 features were used to train a linear SVM classification model. The testing data points were fed into the fine-tuned deep structure to obtain their new feature representations and subsequently were classified by the trained linear SVM classifier. Again, each pilot had an individual model.

Two experiments were designed for comparison. The first one used all 312 EEG features as inputs and the second one utilized principal component analysis (PCA) to reduce the dimensionality of the original EEG features to 30. Both models used linear SVM for classification. In addition, we studied the effect of the hyperparameters in this evaluation.

## IV. RESULTS AND DISCUSSION

In this section, we first discuss the effects of hyperparameters in the deep models in the context of fivefold CV, where the labeled data were randomly divided for training and testing. We then present engagement assessment performances conducted on the continuous labeled data as described in Section III-C.

### A. Results of Fivefold Cross Validation With Different Hyperparameters

*1) Effects of Momentum and Learning Rate in Pretraining:* A good combination of learning rate and momentum is crucial for the convergence of RBM learning. Usually, a small value for the learning rate and a large value for the momentum are helpful for the convergence. Fig. 4 shows reconstruction errors for a Gaussian–Bernoulli RBM with a structure of 312-600. We set the momentum value as 0.9 and the learning rate values as 0.005, 0.01, and 0.018, respectively. It can be observed that the error decreased faster if a larger learning rate was used. However, when we increased the learning rate to 0.02, the training failed to converge. We also tried different values for momentum and found that the training failed to converge when the momentum value was less than 0.6 with a learning rate of 0.018. To guarantee a safe convergence of the RBM training, we set the momentum as 0.9 and the learning rate as 0.01 in subsequent experiments.

*2) Effect of the Deep Network Structure:* We conducted a fivefold CV on the available dataset using five different structures including

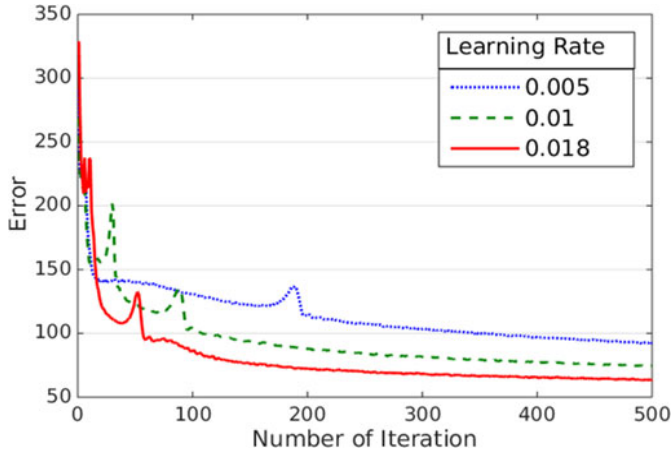

Fig. 4. Learning error for different learning rates.

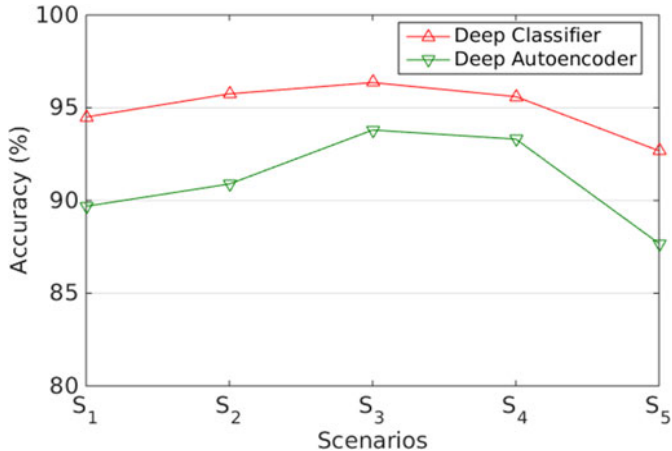

Fig. 5. Accuracies of networks with different structures. ($S_1$ : 600-200 – 100-20, $S_2$ : 200-100-20, $S_3$ : 100-20, $S_4$ : 20, $S_5$ : 800-200-100-10).

TABLE II
RESULTS FOR DROPOUT

|  | Scenario 1 | Scenario 2 | Scenario 3 |
|---|---|---|---|
| Deep Classifier | 97.53 (0.13) | 97.43 (0.17) | 97.16 (0.04) |
| Deep Autoencoder | 91.42 (0.21) | 93.60 (0.26) | 93.86 (0.22) |

Scenario 1: Fine-tuning without dropout. Scenario 2: Fine-tuning with dropout probabilities for visible/hidden layers of 0/0.5. Scenario 3: Fine-tuning with dropout probabilities for visible/hidden layers of 0.2/0.5.

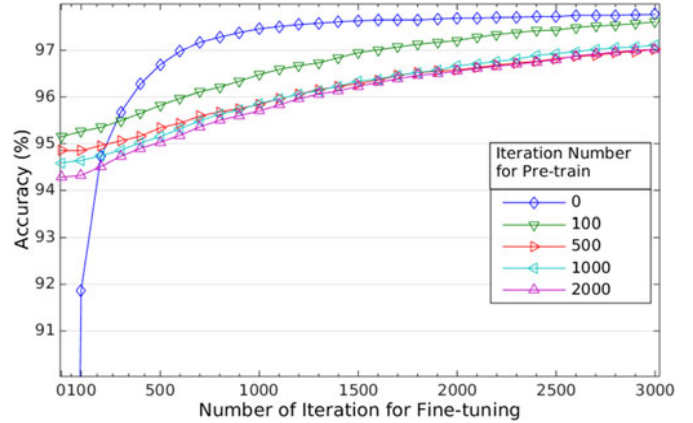

Fig. 6. Performance with different number of iterations (deep classifier).

600-200-100-20, 200-100-20, 100-20, 20, and 800-200-100-10. For each structure, we first pretrained the network, and then, the network was fine-tuned as a deep classifier and a deep autoencoder, respectively. The number of pretraining iterations for all networks was set as 3000 for a fair comparison. Each experiment was repeated five times, and the average accuracy was computed as shown in Fig. 5. It was observed that the highest accuracy of 96.36% was achieved by the network that had a structure of 100-20 and was fine-tuned by labels (deep classifier). Based on this result, the structure of 100-20 was chosen for subsequent engagement assessment.

*3) Effect of Dropout:* We tested if the dropout technique can improve the classification performance. We designed three scenarios for the network with the structure of 100-20: *Scenario 1*: fine-tuning without dropout; *Scenario 2*: fine-tuning with dropout probabilities for visible/hidden layers of 0/0.5; and *Scenario 3*: fine-tuning with dropout probabilities for visible/hidden layers of 0.2/0.5. Note that for each scenario, we repeated the experiment on each subject separately for five times, and the overall average and standard deviation were calculated for each scenario across all subjects.

Table II illustrates the classification performances, where the numbers shown in the table are averaged accuracies from five runs and their corresponding standard deviations are shown in parentheses. For the deep classifier, the performance differences among these three scenarios are very small. However, for the deep autoencoder, the model without dropout was outperformed by the other two models with dropout and the model with dropout probabilities of 0.2/0.5 performed the best. Therefore, dropout with probability of 0.2/0.5 was utilized for subsequent experiments.

*4) Effect of Number of Iterations in Pretraining and Fine-Tuning:* A model trained with a large number of iterations does not always perform better than those with less training because a model could be overfitted [37], [38]. We tried different numbers of iterations in pretraining (0, 100, 500, 1000, and 2000) and fine-tuning (0–3000) and monitored the performances of the model. Results are shown in Fig. 6 (deep classifier) and Fig. 7 (deep autoencoder). Other hyperparameters in the learning were set as follows: momentum = 0.9, learning rate = 0.01, and dropout probability = 0.2/0.5.

If the number of fine-tuning iterations was zero, the networks were initialized with pretrained weights without fine-tuning. Note that the pretraining was unsupervised and label information was not utilized by the model. However, the features learned from pretraining could already achieve over 94% accuracy by both the deep classifier and the deep autoencoder. As a comparison, both models obtained around 50% accuracy, equivalent to a random guess, if models were not pretrained and not fine-tuned. Although pretraining played a positive role, increasing iterations of pretraining did not always result in better results. In this study, 100 iterations of pretraining can achieve good performances.

With the increase of fine-tuning iterations, all deep classifier models kept improving, and the performance of pretrained autoencoders remained the same or dropped slightly. However, there was no trend of overfitting. We believe this is because of the regularization effect of the dropout technique that can prevent weights from coadaptation [22].

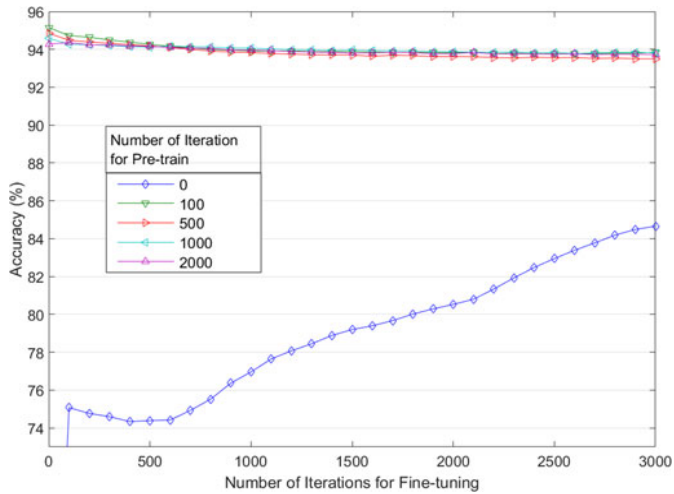

Fig. 7. Performance with different number of iterations (deep autoencoder).

It is also observed that pretraining has a large effect on the deep autoencoder. Fig. 7 shows that the deep autoencoder without pretraining did not perform well even with an increased number of fine-tunings. However, the deep classifier seemed to depend less on pretraining; fine-tuning could improve the model even it was not pretrained (see Fig. 6). The reason could be that the deep classifier had a relatively simple structure ([312-100-20-2]), and also in fivefold CV, the training dataset was relatively larger than those in the experiments we performed in Section IV-B. However, for the deep autoencoder, the network structure was more complicated ([312-100-20-100-312]), and pretraining became a key procedure to improve performance.

### B. Results of Engagement Assessment With Scarce Label Information

*1) Model Comparison Results:* We are more interested in engagement assessment when there are only limited labeled data available and the model performance on sequential data. We compared several models for this purpose including a linear SVM using all 312 raw features (Method 1), a linear SVM using top 30 principal components of the raw features (Method 2), deep classifier (Method 6), deep autoencoder (Method 10), and the two deep models with different combinations with pretraining and dropout (Methods 3–5 and Methods 7–9, respectively). The results are illustrated in Table III and Fig. 8. Other hyperparameters of the deep learning were set as follows: momentum = 0.90, learning rate = 0.01, network structure = [100-20], and the dropout probabilities were chosen as 0.2/0.5. Each experiment except those with Methods 1 and 2 were repeated five times and average accuracies of the five runs are shown in Table III. The standard deviations of the average accuracies are shown in parentheses.

It is observed that both deep learning models outperformed the two compared models, especially when there were not enough labeled data for training (less than 15% of the labeled data for training). The two deep models performed similarly in all of the experiments. These experimental results show that deep models are especially suitable for data modeling in the situation where label data are difficult to obtain.

It also can be observed in Table III and Fig. 8 that the deep models' performances dropped significantly if the models were not pretrained or not trained by the dropout technique when labeled data were limited. With more labeled data for training (see the last column in Table III), the deep classifier performed relatively well even without pretraining or dropout. On the contrary, the deep autoencoder still could not perform well for this case if the deep structure was not pretrained or trained without the dropout technique.

TABLE III
ENGAGEMENT ASSESSMENT RESULTS

| | Top 1% | Top 3% | Top 5% | Top 10% | Top 15% | Top 20% |
|---|---|---|---|---|---|---|
| Method 1 low level features | 62.73 | 67.19 | 73.38 | 79.18 | 81.47 | 84.92 |
| Method 2 PCA features | 58.21 | 63.35 | 67.94 | 71.65 | 73.86 | 77.39 |
| Method 3 Deep classifier + pretraining + dropout | 77.07 (1.02) | **80.45** (**1.31**) | 82.82 (1.21) | 85.22 (1.67) | **85.78** (**0.64**) | **86.52** (**0.72**) |
| Method 4 Deep classifier - pretraining + dropout | 74.66 (1.43) | 78.6 (2.38) | 80.54 (1.22) | 83.68 (0.59) | 85.35 (0.26) | 85.34 (0.79) |
| Method 5 Deep classifier + pretraining - dropout | 70.81 (2.12) | 76.54 (1.28) | 79.45 (1.62) | 84.27 (0.76) | 84.37 (0.84) | 86.00 (0.70) |
| Method 6 Deep classifier - pretraining - dropout | 72.53 (1.74) | 76.08 (2.43) | 79.32 (1.98) | 83.56 (0.64) | 83.76 (0.72) | 85.17 (1.02) |
| Method 7 Deep autoencoder + pretraining + dropout | **77.09** (**0.59**) | 79.54 (0.30) | **83.32** (**0.58**) | **85.74** (**0.39**) | 84.8 (0.63) | 84.83 (0.44) |
| Method 8 Deep autoencoder - pretraining + dropout | 72.49 (0.93) | 75.72 (0.15) | 76.94 (1.00) | 79.53 (0.91) | 78.62 (0.43) | 79.21 (0.25) |
| Method 9 Deep autoencoder + pretraining - dropout | 71.28 (1.76) | 75.77 (0.40) | 75.43 (0.50) | 75.61 (0.40) | 76.32 (0.28) | 77.03 (0.22) |
| Method 10 Deep autoencoder - pretraining - dropout | 73.24 (1.36) | 74.13 (0.71) | 74.59 (0.45) | 75.22 (0.38) | 74.53 (0.23) | 75.53 (0.86) |

Note that "+" and "-" mean w/o a technique.

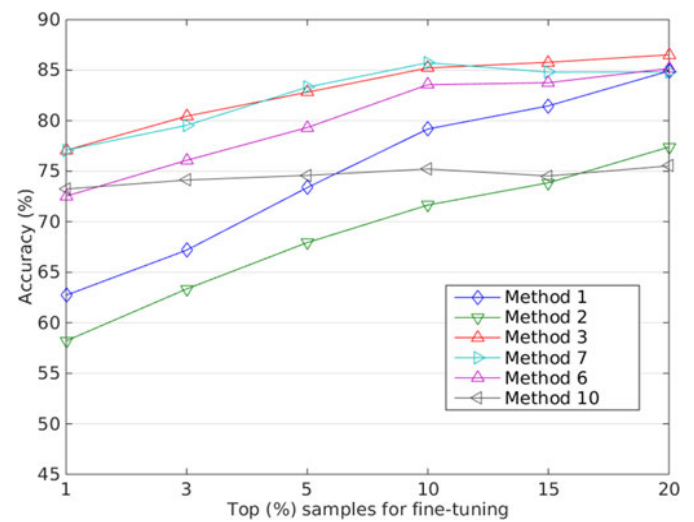

Fig. 8. Engagement assessment results.

The above results are not surprising because the pretraining step in the two deep models may help classification (modeling $p(y|x)$) by modeling $p(x)$ first [39]. In other words, if data labels are limited, understanding the data itself may be important in data classification. Additionally, the dropout technique is a method for mitigating the overfitting problem, and it is especially helpful if labeled data are limited.

A deep classifier usually has less parameters than a deep autoencoder and is preferred in applications where labeled data are difficult to collect. We also showed that dropout and pretraining are both helpful for deep learning with scarce label information especially if the deep model contains a large number of parameters. We should always consider these two regularization techniques for real applications.

## C. Comparison With the Literature

Our results of the deep models in fivefold CV ($\sim 97\%$) are comparable with other functional state assessment studies in the literature. For example, using a multiclass SVM classifier based on EEG signals, Shen *et al.* achieved a tenfold CV accuracy of 91.2% for fatigue modeling [40]. In [29], EEG-based models for mental fatigue obtained fivefold CV accuracies ranging from 90% to 100% with a mean of 97% to 98%. In [41], an EEG-based cognitive state estimation system achieved an accuracy of 98%. However, the more strict evaluation using continuous data blocks for training and testing in our study showed that the CV accuracies of deep models dropped from $\sim 97\%$ to $\sim 85\%$. The more strict evaluation scheme is similar to the real application scenario. We must be aware of this performance drop if we want to deploy such systems for real applications.

We have not found studies similar to ours in the literature that uses scarce label information for engagement assessment. Our study is also an extension to the application of deep learning. While ordinary applications of deep learning (image classification, speech recognition, etc.) utilize large datasets for training, we successfully extended deep learning models to small-sized data in this study.

## D. Computational Efficiency

The running time for deep learning algorithms depends on the number of training samples, the structure of the network, and the number of iterations. Once a model has been trained, it requires almost no time for testing. For our cases, the deep learning algorithms were used for feature extraction, and linear SVM models were trained for classification so that the running time consisted of time for training the deep networks and the SVM models. All computations were done on an HP-Z800 workstation, and the hardware configuration includes two Intel Xeon x5660, 48-G memory, and a GeForce GTX 780 GPU with 3-G RAM. The program was developed in MATLAB and accelerated with the GPU. For a typical scenario with 1200 data samples, a network structure of 100-20, the number of iterations set to 2000 for both pretraining and fine-tuning, and a linear SVM as the classification model, the running time for one subject was around 10 min for either the deep classifier or the deep autoencoder. The running time was increased by a factor of 5 for fivefold CV.

## E. Limitations of the Study

Our study has limitations. First, we did not study the physiological significance of the learned high-level representations. To analyze the significance, we will need to visualize the feature hierarchies learned by the deep models. A feature learned by a hidden unit is defined as a particular signal pattern at input that maximizes the output of the hidden unit. This pattern can be achieved by maximizing the hidden unit's output with respect to the input of the deep models, which currently is under investigation. The patterns will be in the input space (PSD features) that will enable us to interpret physiological meaning of the deep models for engagement assessment. Second, we did not address the individual variations among subjects. In the two evaluation experiments, each pilot's model was trained by his/her own data. We tested these individual models across subject and the performance dropped significantly. We have developed systematic methods to handle the issue, and a separate paper is under preparation.

## V. CONCLUSION

This paper has studied engagement assessment of pilots under complex tasks with limited labeled EEG data. We developed deep learning models that were able to learn valuable high-level features by taking advantage of both unlabeled and labeled data. Two deep models (a deep classifier and a deep autoencoder) have been studied, and both models outperformed two traditional methods when label information was limited for training. Pretraining and dropout are two regularization techniques, and both were found beneficial in deep learning for engagement assessment. We also showed that deep classifier is preferred as compared with deep autoencoder because the former usually has less parameters to optimize.